\documentclass[preprint,12pt]{elsarticle}

\usepackage{amssymb}

\usepackage{natbib}
\usepackage{caption, subcaption}
\usepackage{graphicx}

\usepackage[section]{placeins}
\journal{Magnetic Resonance Imaging}

\begin{document}

\begin{frontmatter}



\title{Machine learning in resting-state fMRI analysis}


 \author[label1]{Meenakshi Khosla}
  \author[label2]{Keith Jamison}
   \author[label1]{Gia H. Ngo}
  \author[label2,label3]{Amy Kuceyeski}
   \author[label1,label4]{Mert R. Sabuncu\footnote{Please address correspondence to: Dr. Mert R. Sabuncu (msabuncu@cornell.edu), School of Electrical and Computer Engineering, 300 Frank H. T. Rhodes Hall,
Cornell University, Ithaca, NY 14853.}}
\address[label1]{School of Electrical and Computer Engineering, Cornell University}
 \address[label2]{Radiology, Weill Cornell Medical College}
\address[label3]{Brain and Mind Research Institute, Weill Cornell Medical College}
\address[label4]{Nancy E. \& Peter C. Meinig School of Biomedical Engineering, Cornell University}

\begin{abstract}
Machine learning techniques have gained prominence for the analysis of resting-state functional Magnetic Resonance Imaging (rs-fMRI) data.  Here, we present an overview of various unsupervised and supervised machine learning applications to rs-fMRI.  We present a methodical taxonomy of machine learning methods in resting-state fMRI. We identify three major divisions of unsupervised learning methods with regard to their applications to rs-fMRI, based on whether they discover principal modes of variation across space, time or population. Next, we survey the algorithms and rs-fMRI feature representations that have driven the success of supervised subject-level predictions. The goal is to provide a high-level overview of the burgeoning field of rs-fMRI from the perspective of machine learning applications. 
\end{abstract}

\begin{keyword}
Machine learning \sep resting-state \sep functional MRI \sep intrinsic networks \sep brain connectivity

\end{keyword}

\end{frontmatter}


\section{Introduction}

Resting-state fMRI (rs-fMRI) is a widely used neuroimaging tool that measures spontaneous fluctuations in neural blood oxygen-level dependent (BOLD) signal across the whole brain, in the absence of any controlled experimental paradigm. 
In their seminal work, Biswal et al. \cite{Biswal1995} demonstrated temporal coherence of low-frequency spontaneous fluctuations between long-range functionally related regions of the primary sensory motor cortices even in the absence of an explicit task, suggesting a neurological significance of resting-state activity. Several subsequent studies similarly reported other collections of regions co-activated by a task (such as language, motor, attention, audio or visual processing etc.) that show correlated fluctuations at rest \cite{Beckmann2005, cordes2000, Damoiseaux2006, DeLuca2005, dosenbach2007distinct, fox2006spontaneous, hampson2002detection, margulies2007mapping, seeley2007dissociable, Smith2009}. These spontaneously co-fluctuating regions came to be known as the resting state networks (RSNs) or intrinsic brain networks. The term RSN henceforth denotes brain networks subserving shared functionality as discovered using rs-fMRI. 


Rs-fMRI has enormous potential to advance our understanding of the brain's functional organization and how it is altered by damage or disease. A major emphasis in the field is on the analysis of resting-state functional connectivity (RSFC) that measures statistical dependence in BOLD fluctuations among spatially distributed brain regions. Disruptions in RSFC have been identified in several neurological and psychiatric disorders, such as Alzheimer's \cite{greicius2004default, supekar2008, sheline2013resting}, autism \cite{kennedy2008intrinsic, monk2009abnormalities, hull2017resting}, depression \cite{anand2005activity, greicius2007resting, mulders2015resting}, schizophrenia \cite{liang2006widespread, sheffield2016cognition}, etc.  Dynamics of RSFC have also garnered considerable attention in the last few years, and a crucial challenge in rs-fMRI is the development of appropriate tools to capture the full extent of this RS activity. rs-fMRI captures a rich repertoire of intrinsic mental states or spontaneous thoughts and, given the necessary tools, has the potential to generate novel neuroscientific insights about the nature of brain disorders \cite{cc2009dep, chen2011alz, Nielsen2013, fox2010clinical, greicius2008resting, zhang2010disease}.

The study of rs-fMRI data is highly interdisciplinary, majorly influenced by fields such as machine learning, signal processing and graph theory. Machine learning methods provide a rich characterization of rs-fMRI, often in a data-driven manner.  Unsupervised learning methods in rs-fMRI are focused primarily on understanding the functional organization of the healthy brain and its dynamics.
For instance, methods such as matrix decomposition or clustering can simultaneously expose multiple functional networks within the brain and also reveal the latent structure of dynamic functional connectivity.

Supervised learning techniques, on the other hand, can harness RSFC to make individual-level predictions. Substantial effort has been devoted to using rs-fMRI for classification of patients versus controls, or to predict disease prognosis and guide treatments. Another class of studies explores the extent to which individual differences in cognitive traits may be predicted by differences in RSFC, yielding promising results. Predictive approaches can also be used to address research questions of interest in neuroscience. For example, is RSFC heritable? Such questions can be formulated within a prediction framework to test novel hypotheses. 

From mapping functional networks to making individual-level predictions, the applications of machine learning in rs-fMRI are far-reaching. The goal of this review is to present in a concise manner the role machine learning has played in generating pioneering insights from rs-fMRI data, and describe the evolution of machine learning applications in rs-fMRI. 
We will present a review of the key ideas and application areas for machine learning in rs-fMRI rather than delving into the precise technical nuances of the machine learning algorithms themselves. In light of the recent developments and burgeoning potential of the field, we discuss current challenges and promising directions for future work.

\subsection{Resting-state fMRI: A Historical Perspective }

Until the 2000s, task-fMRI was the predominant neuroimaging tool to explore the functions of different brain regions and how they coordinate to create diverse mental representations of cognitive functions. The discovery of correlated spontaneous fluctuations within known cortical networks by Biswal et al. \cite{Biswal1995} and a plethora of follow-up studies have established rs-fMRI as a useful tool to explore the brain's functional architecture. 
Studies adopting the resting-state paradigm have grown at an unprecedented scale over the last decade. These are much simpler protocols than alternate task-based experiments, capable of providing critical insights into functional connectivity of the healthy brain as well as its disruptions in disease. 
Resting-state is also attractive as it allows multi-site collaborations, unlike task-fMRI that is prone to confounds induced by local experimental settings. This has enabled network analysis at an unparalleled scale. 



Traditionally, rs-fMRI studies have focused on identifying spatially-distinct yet functionally associated brain regions through seed-based analysis (SBA). In this approach, seed voxels or regions of interest are selected \textit{a priori} and the time series from each seed is correlated with the time series from all brain voxels to generate a series of correlation maps. SBA, while simple and easily interpretable, is limited since it is heavily dictated by manual seed selection and, in its simplest form, can only reveal one specific functional system at a time.


Decomposition methods like Independent Component Analysis (ICA) emerged as a highly promising alternative to seed-based correlation analysis in the early 2000s \cite{cordes, Beckmann2005, Beckmann04}. 
This was followed by other unsupervised learning techniques such as clustering. 
In contrast to seed-based methods that explore networks associated with a seed voxel (such as motor or visual functional connectivity maps), these new class of model-free methods based on decomposition or clustering explored RSNs simultaneously across the whole brain for individual or group-level analysis. Regardless of the analysis tool, all studies largely converged in reporting multiple robust resting-state networks across the brain, such as the primary sensorimotor network, the primary visual network, fronto-parietal attention networks and the well-studied default mode network. Regions in the default mode network, such as the posterior cingulate cortex, precuneus, ventral and dorsal medial prefrontal cortex, show increased levels of activity during rest than tasks suggesting that this network represents the baseline or default functioning of the human brain. The default mode network has sparked a lot of interest in the rs-fMRI community \cite{Greicius2003dmn}, and several studies have consequently explored disruptions in DMN resting-state connectivity in various neurological and psychiatric disorders, including autism, schizophrenia and Alzheimer's. \cite{jung2014, ongur2010,koch2012} 

Despite the widespread success and popularity of rs-fMRI, the causal origins of ongoing spontaneous fluctuations in the resting brain remain largely unknown. Several studies explored whether resting-state coherent fluctuations have a neuronal origin, or are just manifestations of aliasing or physiological artifacts introduced by the cardiac or respiratory cycle. Over time, evidence in support for a neuronal basis of BOLD-based resting state functional connectivity has accumulated from multiple complementary sources. This includes (a) observed reproducibility of RSFC patterns across independent subject cohorts \cite{DeLuca2005, Damoiseaux2006}, (b) its persistence in the absence of aliasing and distinct separability from noise components \cite{DeLuca2005, Cordes2001noise}, (c)  its similarity to known functional networks \cite{Biswal1995, Beckmann2005, Smith2009} and (d) consistency with anatomy \cite{Salvador2005, Mezer2009}, (e) its correlation with cortical activity studied using other modalities \cite{Laufs2003, Damoiseaux2009, Nir2008} and finally, (f) its systematic alterations in disease \cite{cc2009dep, chen2011alz,Nielsen2013}. 

\begin{figure}
\hspace*{-0.8cm}
\centering
  \includegraphics[width=0.56\textwidth,keepaspectratio]{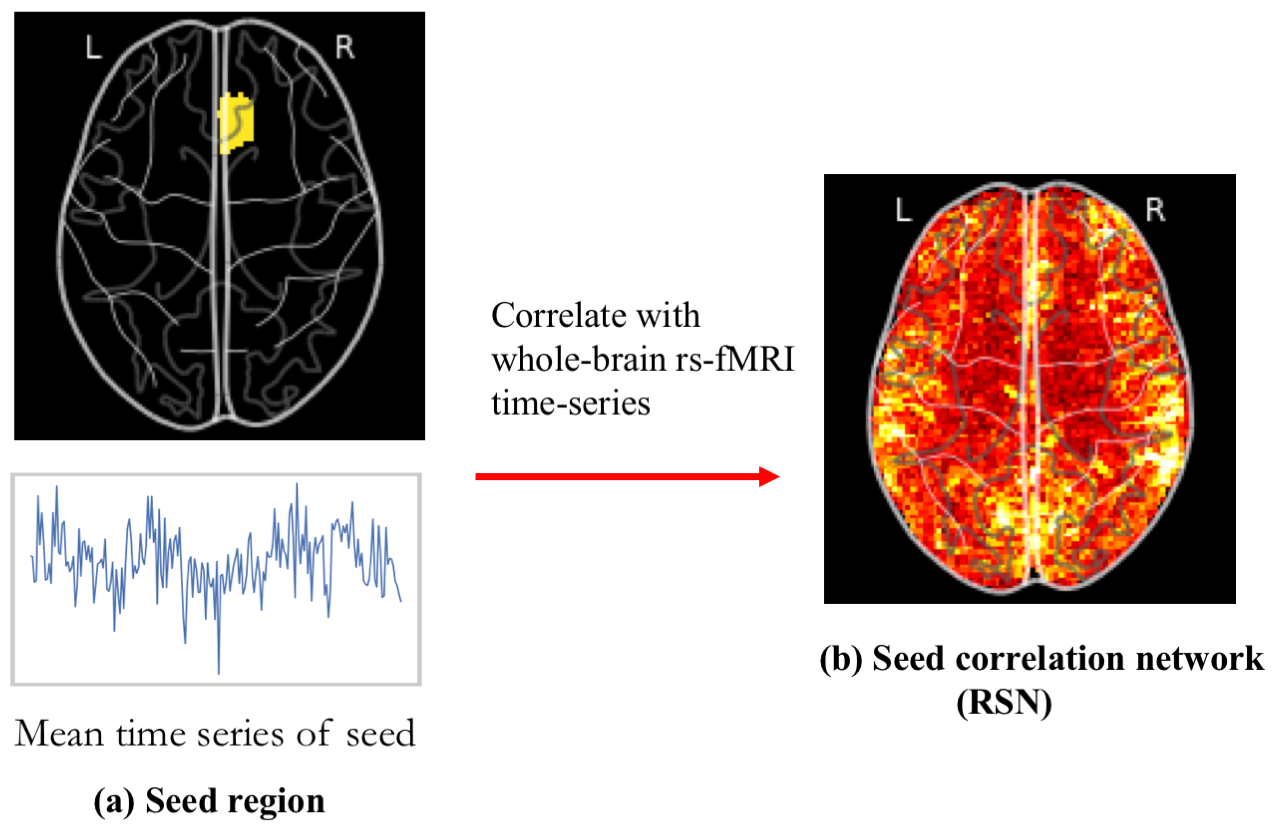}
  \caption{Traditional seed based analysis approach}
  \label{fig:seed}
\end{figure}

\subsection{Application of Machine Learning in rs-fMRI}

A vast majority of literature on machine learning for rs-fMRI is devoted to unsupervised learning approaches. Unlike task-driven studies, modelling resting-state activity is not straightforward since there is no controlled stimuli driving these fluctuations. Hence, analysis methods used for characterizing the spatio-temporal patterns observed in task-based fMRI are generally not suited for rs-fMRI. Given the high dimensional nature of fMRI data, it is unsurprising that early analytic approaches focused on decomposition or clustering techniques to gain a better characterization of data in spatial and temporal domains. Unsupervised learning approaches like ICA catalyzed the discovery of the so-called resting-state networks or RSNs. Subsequently, the field of resting-state brain mapping expanded with the primary goal of creating brain parcellations, i.e., optimal groupings of voxels (or vertices in the case of surface representation) that describe functionally coherent spatial compartments within the brain. These parcellations aid in the understanding of human functional organization by providing a reference map of areas for exploring the brain's connectivity and function. Additionally, they serve as a popular data reduction technique for statistical analysis or supervised machine learning. 

More recently, departing from the stationary representation of brain networks, studies have shown that RSFC exhibits meaningful variations during the course of a typical rs-fMRI scan \cite{chang2010dynamic, Allen2014}. 
Since brain activity during resting-state is largely uncontrolled, this makes network dynamics even more interesting. 
Using unsupervised pattern discovery methods, resting-state patterns have been shown to transition between  discrete recurring functional connectivity "states", representing diverse mental processes \cite{Allen2014, Vidaurre2017, Eavani2013}. In the simplest and most common scenario, dynamic functional connectivity is expressed using sliding-window correlations. In this approach, functional connectivity is estimated in a temporal window of fixed length, which is subsequently shifted by different time steps to yield a sequence of correlation matrices. Recurring correlation patterns can then be identified from this sequence through decomposition or clustering. This dynamic nature of functional connectivity opens new avenues for understanding the flexibility of different connections within the brain as they relate to behavioral dynamics, with potential clinical utility \cite{Reinen2018}. 

Another, perhaps clinically more promising application of machine learning in rs-fMRI expanded in the late 2000s. This new class of applications leveraged supervised machine learning for individual level predictions. The covariance structure of resting-state activity, more popularly known as the "connectome", has garnered significant interest in the field of neuroscience as a sensitive biomarker of disease.
Studies have further shown that an individual's connectome is unique and reliable, akin to a fingerprint \cite{Finn2015}. 
Machine learning can exploit these neuroimaging based biomarkers to build diagnostic or prognostic tools. 
Visualization and interpretation of these models can complement statistical analysis to provide novel insights into the dysfunction of resting-state patterns in brain disorders. 
Given the prominence of deep learning in today's era, several novel neural-network based approaches have also emerged for the analysis of rs-fMRI data. 
A majority of these approaches target connectomic feature extraction for single-subject level predictions.

In order to organize the work in this rapidly growing field, we sub-divide the machine learning approaches into different classes by methods and application focus. 
We first differentiate among unsupervised learning approaches based on whether their main focus is to discover (a) the underlying spatial organization that is reflected in coherent fluctuations, (b) the  structure in temporal dynamics of resting-state connectivity,  or (c) population-level structure for inter-subject comparisons. 
Next, we move on to discuss supervised learning. We organize this section by discussing the relevant rs-fMRI features employed in these models, followed by commonly used training algorithms, and finally the various application areas where rs-fMRI has shown promise in performing predictions. 

\section{Unsupervised Learning}
The primary objective of unsupervised learning is to discover latent representations and disentangle the explanatory factors for variation in rich, unlabelled data. 
These learning methods do not receive any kind of supervision in the form of target outputs (or labels) to guide the learning process. 
Instead, they focus on learning structure in the data in order to extract relevant signal from noise. 
Unsupervised machine learning methods have proven promising for the analysis of high-dimensional data with complex structures, making it ever more relevant to rs-fMRI. 

Many unsupervised learning approaches in rs-fMRI aim to parcellate the brain into discrete functional sub-units, akin to atlases. 
These segmentations are driven by functional data, unlike those approaches that use cytoarchitecture as in the Broadmann atlas, or macroscopic anatomical features, as in the  Automated Anatomical Labelling (AAL) atlas~\cite{AAL2002}.  
A second class of applications delve into the exploration of brain network dynamics. Unsupervised learning has recently been applied to interrogate the dynamic functional connectome with promising results\cite{Vidaurre2017, Allen2014, Eavani2013, Leonardi2013, Leonardi2014}. Finally, the third application of unsupervised learning focuses on learning latent low-dimensional representations of RSFC to conduct analyses across a population of subjects. We discuss the methods under each of these challenging application areas below.  




\begin{figure}
\centering
  \includegraphics[width=9cm,keepaspectratio]{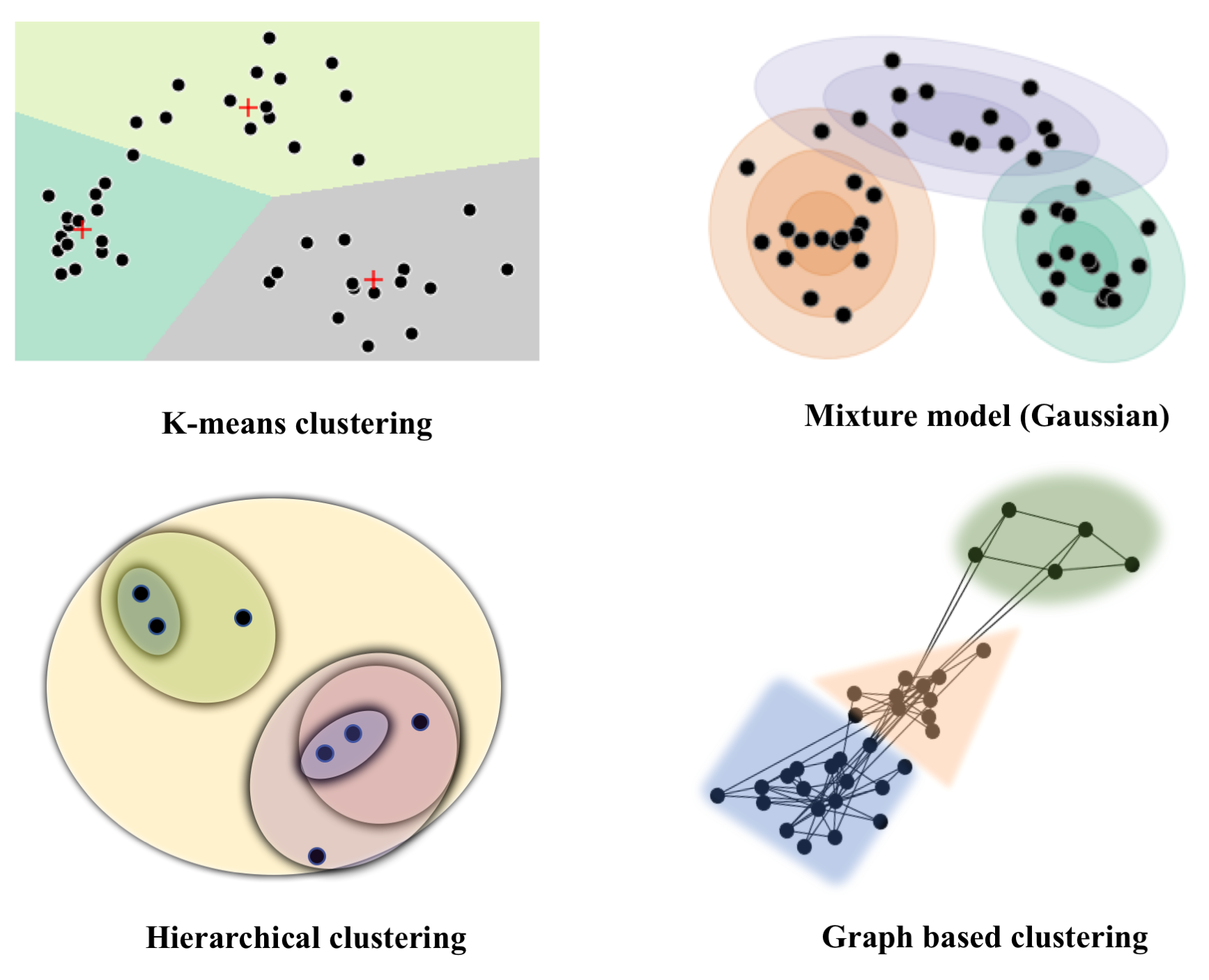}
  \caption{Illustrations of popular clustering algorithms: K-means clustering partitions the data space into Voronoi cells,  where each observation is assigned to the cluster with the nearest centroid (marked red in the figure). GMMs assume that each cluster is sampled from a multivariate Gaussian distribution and estimates these probability densities to generate probabilistic assignment of observations to different clusters. Hierarchical (agglomerative) clustering generates nested partitions, where partitions are merged iteratively based on a linkage criteria. Graph-based clustering partitions the graph representation of data so that, for example, number of edges connecting distinct clusters are minimal.}
\end{figure}

\subsection{Discovering spatial patterns with coherent fluctuations\label{app1}}
Mapping the boundaries of functionally distinct neuroanatomical structures, or identifying clusters of functionally coupled regions in the brain is a major objective in neuroscience. Rs-fMRI and machine learning methods provide a promising combination with which to achieve this loft goal.


Decomposition or factorization based approaches assume that the observed data can be decomposed as a product of simpler matrices, often imposing a specific structure and/or sparsity on these individual matrices. 
In the case of rs-fMRI, the typical approach is to decompose the 4D fMRI data into a linear superposition of distinct spatial modes that show coherent temporal dynamics. 


\begin{figure*}
\hspace*{-1cm}
\centering
  \includegraphics[width=14cm,keepaspectratio]{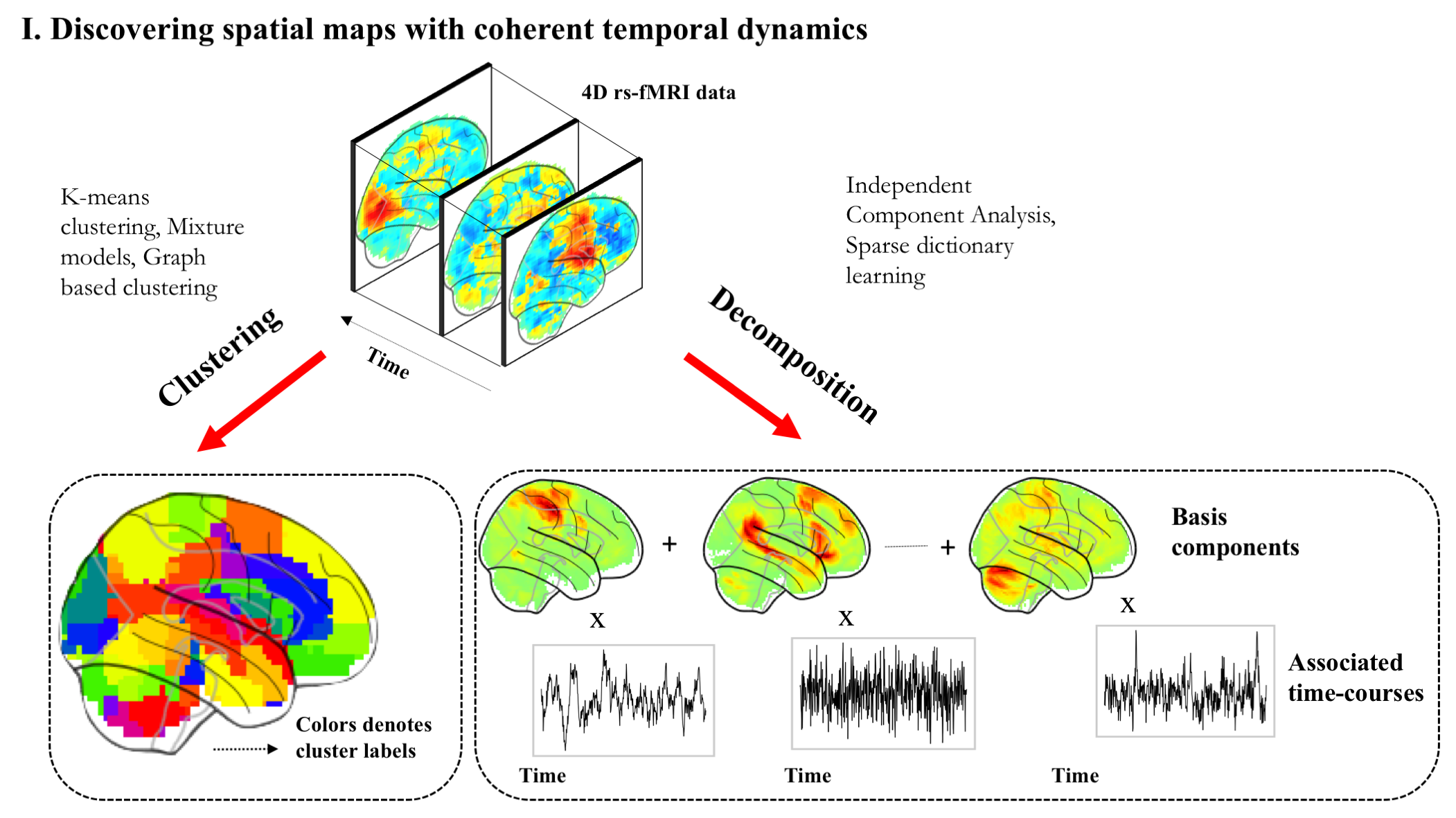}
  \caption{Schematic of application \ref{app1}: In decomposition, the original fMRI data is expressed as a linear combination of spatial patterns and their associated time series - in ICA, the independence of spatial maps is optimized whereas in sparse dictionary learning, the sparsity of maps is encouraged. In clustering, time series or connectivity fingerprints of voxels are clustered to assign voxels to distinct functional networks.}
\end{figure*}

\subsubsection{ICA} 

Independent Component Analysis (ICA) is a popular decomposition method for data that assumes a linear combination of statistically independent sources. Independent components (ICs) are usually computed by either minimizing the mutual information between sources (InfoMax) or by maximizing their non-gaussianity (FastICA). 
ICA has been one of the earliest and most widely used analytic tools for rs-fMRI, driving several pivotal neuroscientific insights into intrinsic brain networks. When applied to rs-fMRI, brain activity is expressed as a linear superposition of distinct spatial patterns or maps, with each map following its own characteristic time course. These spatial maps can reflect a coherent functional system or noise, and several criteria can be used to automatically differentiate them. This capability to isolate noise sources makes ICA particularly attractive. In the early days of rs-fMRI, several studies demonstrated marked resemblance between the ICA spatial maps and cortical functional networks known from task-activation studies \cite{Beckmann2005,Damoiseaux2006}. 
While typical ICA models are noise-free and assume that the only stochasticity is in the sources themselves, several variants of ICA have been proposed to model additive noise in the observed signals. 
Beckmann et al. \cite{Beckmann2005} introduced a probabilistic ICA (PICA) model to extract the connectivity structure of rs-fMRI data. PICA models a linear instantaneous mixing process under additive noise corruption and statistical independence between sources. 
De Luca et al. \cite{DeLuca2005} showed that PICA can reliably distinguish RSNs from artifactual patterns. Both these works showed high consistency in resting-state patterns across multiple subjects. While there is no standard criteria for validating the ICA patterns, or any clustering algorithm for that matter, reproducibility or reliability is often used for quantitative assessment. 
   
ICA can also be extended to make group inferences in population studies. Group ICA is thus far the most widely used strategy, where multi-subject fMRI data are concatenated along the temporal dimension before implementing ICA \cite{pmid11559959}. Individual-level ICA maps can then be obtained from this group decomposition by back-projecting the group mixing matrix \cite{pmid11559959}, or using a dual regression approach \cite{beckmann2009dual}. More recently, Du et al.\cite{Du2013} introduced a group information guided ICA to preserve statistical independence of individual ICs, where group ICs are used to constrain the corresponding subject-level ICs. Varoquaux et al. \cite{Varoquaux2010a} proposed a robust group-level ICA model to facilitate between-group comparisons of ICs. They introduce a generative framework to model two levels of variance in the ICA patterns, at the group level and at a subject-level, akin to a multivariate version of mixed-effect models. The IC estimation procedure, termed Canonical ICA, employs Canonical Correlation Analysis to identify a joint subspace of common IC patterns across subjects and yields ICs that are well representative of the group.  


Alternatively, it is also possible to compute individual-specific ICA maps and then establish correspondences across them \cite{Du2013} for generating group inferences; however, this approach has been limited because source separations can be very different across subjects, for example, due to fragmentation.

While ICA and its extensions have been used broadly by the rs-fMRI community, it is important to acknowledge its limitations. ICA models linear representations of non-Gaussian data. Whether a linear transformation can adequately capture the relationship between independent latent sources and the observed high-dimensional fMRI data is uncertain and likely unrealistic. Unlike the popular Principal Component Analysis (PCA), ICA does not provide the ordering or the energies of its components, which makes it impossible to distinguish strong and weak sources. This also complicates replicability analysis since known sources i.e., spatial maps could be expressed in any arbitrary order. Extracting meaningful ICs also sometimes necessitates manual selection procedures, which can be inefficient or subjective. In the ideal scenario, each individual component represents either a physiologically meaningful activation pattern or noise. However, this might be an unrealistic assumption for rs-fMRI. Additionally, since ICA assumes non-Gaussianity of sources, Gaussian physiological noise can contaminate the extracted components. Further, due to the high-dimensionality of fMRI, analysis often proceeds with PCA based dimensionality reduction before application of ICA. PCA computes uncorrelated linear transformations of highest variance (thus explaining greatest variability within the data) from the top eigenvectors of the data covariance matrix. While this step is useful to remove observation noise, it could also result in loss of signal information that might be crucial for subsequent analysis. Although ICA optimizes for independence, it does not guarantee independence. Based on studies of functional integration within the brain, assumptions of independence between functional units could themselves be questioned from a neuroscientific point of view. 
Several papers have suggested that ICA is especially effective when spatial patterns are sparse, with negligible or little overlap. 
This hints to the possibility that success of ICA is driven by sparsity of the components rather than their independence.
Along these lines, Daubechies and colleagues claim that fMRI representations that optimize for sparsity in spatial patterns are more effective than fMRI representations that optimize independence~\cite{Daubechies10415}.

\subsubsection{Learning sparse spatial maps} 

Sparse dictionary learning is another popular framework for constructing succinct representations of observed data. 
Dictionary learning is formulated as a linear decomposition problem, similar to ICA/PCA, but with sparsity constraints on the components.  


Varoquaux et al. \cite{varoquaux2011msdl} adopt a dictionary learning framework for segmenting functional regions from resting-state fMRI time series. Their approach accounts for inter-subject variability in functional boundaries by allowing the subject-specific spatial maps to differ from the population-level atlas. Concretely, they optimize a loss function comprising a residual term that measures the approximation error between data and its factorization, a cost term penalizing large deviations of individual subject spatial maps from group level latent maps, and a regularization term promoting sparsity.  In addition to sparsity, they also impose a smoothness constraint so that the dominant patterns in each dictionary are spatially contiguous to construct a well-defined parcellation. 
In order to prevent blurred edges caused due to the smoothness constraint, Abraham et al. \cite{abraham2013tv} propose a total variation regularization within this multi-subject dictionary learning framework. This approach is shown to yield more structured parcellations that outperform competing methods like ICA and clustering in explaining test data.  
Similarly, Lv et al. \cite{lv2015} propose a strategy to learn sparse representations of whole-brain fMRI signals in individual subjects by factorizing the time-series into a basis dictionary and its corresponding sparse coefficients. Here, dictionaries represent the co-activation patterns of functional networks and coefficients represent the associated spatial maps. Experiments revealed a high degree of spatial overlap in the extracted functional networks in contrast to ICA that is known to yield spatially non-overlapping components in practice.


Clustering is an alternative unsupervised learning approach for analysis of rs- fMRI data. 
Unlike ICA or dictionary learning, clustering is used to partition the brain surface (or volume) into \textit{disjoint} functional networks.  

It is important to draw a distinction at this stage between two slightly different applications of clustering since they sometimes warrant different constraints; one direction is focused on identifying functional networks which are often spatially distributed, whereas the other is used to parcellate brain regions. 
The latter application aims to construct atlases that reflect local areas that constitute the functional neuroanatomy, much like how standard atlases such as the Automated Anatomical Labelling (AAL) \cite{AAL2002} delineate macroscopic anatomical regions.

One important design decision in the application of clustering is the distance function used to measure dissimilarity between different voxels (or vertices). In the case of rs-fMRI, this distance function is either computed on raw time-series at voxels or between their connectivity profiles. While these two distances are motivated by the same idea of functional coherence, certain differences have been found in parcellations optimized using either criteria \cite{Craddock2012a}. 

An important requirement for almost all of these methods is the \textit{a priori} selection of the number of clusters. These are often determined through cross-validation or through statistics that reflect the quality or stability of partitions at different scales.

\subsubsection{K-means clustering and mixture models} 
K-means clustering is thus far the most popular learning algorithm for partitioning data into disjoint groups. It begins with initial estimates of cluster centroids and iteratively refines them by (a) assigning each datum to its closest cluster, and (b) updating cluster centroids based on these new assignments. This method is frequently used for spatial segmentation of fMRI data \cite{Golland2008,Mezer2009,Lee2012anatomy,kim2010clustering}. Similarity between voxels can be defined by correlating their raw time-series~\cite{Golland2008} or connectivity profiles \cite{kim2010clustering}. Euclidean distance metrics have also been used on spectral features of time series \cite{Mezer2009}. 

K-means clustering has provided several novel insights into functional organization of the human brain. It has revealed the natural division of cortex into two complementary systems, the internally-driven "intrinsic" system and the stimuli-driven "extrinsic" system~\cite{Golland2008, Lee2012anatomy}; provided evidence for a hierarchical organization of RSNs \cite{Lee2012anatomy}; and exposed the anatomical contributions to co-varying resting-state fluctuations \cite{Mezer2009}. 

Mixture models are often used to represent probability densities of complex multimodal data with hidden components. These models are constructed as mixtures of arbitrary unimodal distributions, each representing a distinct cluster. Parameters of this model, which reveal the cluster identities and underlying distributions, are usually optimized within the Expectation-Maximization (EM) framework.

Golland et al. \cite{Golland} proposed a Gaussian mixture model for clustering fMRI signals. Here, the signal at each voxel is modelled as a weighted sum of N Gaussian densities, with N determining the number of hypothesized functional networks and weights reflecting the probability of assignment to different networks. Large-scale systems were explored at several resolutions, revealing an intrinsic hierarchy in functional organization. Yeo et al. \cite{ThomasYeo2011} used rs-fMRI measurements on ~1000 subjects to estimate the organization of large-scale distributed cortical networks. They employed a mixture model to identify clusters of voxels with similar corticocortical connectivity profiles. Number of clusters were chosen from stability analysis and parcellations at both a coarse resolution of 7 networks and a finer scale of 17 networks were identified. A high degree of replicability was attained across data samples, suggesting that these networks can serve as reliable reference maps for functional characterization.

\subsubsection{Identifying hierarchical spatial organization}
Hierarchical clustering methods group the data into a set of nested partitions. This multi-resolution structure is often represented with a cluster tree, or dendrogram. Hierarchical clustering is divided into agglomerative or divisive methods, based on whether the clusters are identified in a bottom-up or top-down fashion respectively. Hierarchical agglomerative clustering (HAC), the more dominant approach for rs-fMRI, iteratively merges the least dissimilar clusters according to a pre-specified distance metric, until the entire data is labelled as one single cluster. 

Many distance metrics have been proposed in literature that optimize the different goals of hierarchical clustering. These include: (a) Single-link, which defines the distance between clusters as the distance between their closest points, (b) complete-link, where this distance is measured between the farthest points, (c) average-link, which measures the average distance between members etc.  Alternate methods for merging have also been proposed, the most popular being Ward's criterion. Ward's method measures how much the within-cluster variance will increase when merging two partitions and minimizes this merging cost. 

Several studies have provided evidence for a hierarchical organization of functional networks in the brain\cite{Golland,Lee2012anatomy}. HAC thus provides a natural tool to partition rs-fMRI data and explore this latent hierarchical structure. Earliest applications of clustering to rsfMRI were based on HAC \cite{cordes2002a, Salvador2005}. This technique thus largely demonstrated the feasibility of clustering for extracting RSNs from rs-fMRI data. Recent applications of HAC have focused on defining whole-brain parcellations for downstream analysis \cite{Blumensath2013, Abraham2017SL, thirion2014}. Spatial continuity can be enforced in parcels, for example, by considering only local neighborhoods as potential candidates for merging \cite{Blumensath2013}. 

An advantage of hierarchical clustering is that unlike k-means clustering, it does not require knowledge of the number of clusters and is completely deterministic. However, once the cluster tree is formed, the dendrogram must be split at a level that best characterizes the "natural" clusters. This can be determined based on a linkage inconsistency criterion \cite{cordes2002a}, consistency across subjects \cite{Salvador2005}, or advance empirical knowledge \cite{wanghac}. 

While a promising approach for rs-fMRI analysis, hierarchical clustering has some inherent limitations. 
It often relies on prior dimensionality reduction, for example by using an anatomical template \cite{Salvador2005}, which can bias the resulting parcellation.  
It is a greedy strategy and erroneous partitions at an early stage cannot be rectified in subsequent iterations. Single-linkage criterion may not work well in practice since it merges partitions based on the nearest neighbor distance, and hence is not inherently robust to noisy resting-state signals. Further, different metrics usually optimize divergent attributes of clusters. For example, single-link clustering encourages extended clusters whereas complete-link clustering promotes compactness. This makes the \textit{a priori} choice of distance metric somewhat arbitrary.

\subsubsection{Graph based clustering} 
Graph based clustering forms another class of similarity-based partitioning methods for data that can be represented using a graph. Given a weighted undirected graph, most graph-partitioning methods, such as the normalized cut (Ncut)~\cite{Shimalik2000}, optimize a dissimilarity measure. Ncut computes the total edge weights connecting two partitions and normalizes this by their weighted connections to all nodes within the graph. Ncut criteria is thus far the most popular graph-based partitioning scheme as it simultaneously minimizes between-cluster similarity while maximizing within-cluster similarity. This clustering approach is also more resilient to outliers, compared to k-means or hierarchical clustering. 

Functional MRI data can be naturally represented in the form of graphs. 
Here, nodes represent voxels and edges represent connection strength, typically measured by a correlation coefficient between voxel time series or between connectivity maps \cite{Craddock2012a, VandenHeuvel2008}. Often, thresholding is applied on edges to limit graph complexity. Graph segmentation approaches have been widely used to derive whole-brain parcellations \cite{Craddock2012a,Shen2013,Honnorat2015}. Population-level parcellations are usually derived in a  two stage procedure: First, individual graphs are clustered to extract functionally-linked regions, followed by a second stage where a group-level graph characterizing the consistency of individual cluster maps is clustered~\cite{VandenHeuvel2008,Craddock2012a}.   
Spatial contiguity can be easily enforced by constraining the connectivity graph to local neighborhoods~\cite{Craddock2012a}, or through the use of shape priors~\cite{Honnorat2015}. Departing from this protocol, Shen et al. \cite{Shen2013} propose a groupwise clustering approach that jointly optimizes individual and group parcellations in a single stage and yields spatially smooth group parcellations in the absence of any explicit constraints. 

A disadvantage of the Ncut criteria for fMRI is its bias towards creating uniformly sized clusters, whereas in reality functional regions show large size variations. Graph construction itself involves arbitrary decisions which can affect clustering performance \cite{maier2011} e.g., selecting a threshold to limit graph edges, or choosing the neighborhood to enforce spatial connectedness. 

\subsubsection{Comments}
\paragraph{I. A comment on alternate connectivity-based parcellations}
Several papers make a distinction between clustering / decomposition and boundary detection based approaches for network segmentation. In the rs-fMRI literature, several non-learning based parcellations have been proposed, that exploit traditional image segmentation algorithms to identify functional areas based on abrupt RSFC transitions \cite{Gordon2016,Glasser2016}.  Clustering algorithms do not mandate spatial contiguity, whereas boundary based methods implicitly do. In contrast, boundary based approaches fail to represent long-range functional associations, and may not yield parcels that are as connectionally homogeneous as unsupervised learning approaches. A hybrid of these approaches can yield better models of brain network organization. This direction was recently explored by Schaefer et al.~\cite{Schaefer2017} with a Markov Random Field model. The resulting parcels showed superior homogeneity compared with several alternate gradient and learning-based schemes. 


\paragraph{II. Subject versus population level parcellations}

Significant effort in rs-fMRI literature is dedicated to identifying population-average parcellations. The underlying assumption is that functional connectivity graphs exhibit similar patterns across subjects, and these global parcellations reflect common organizational principles. Yet, individual-level parcellations can potentially yield more sensitive connectivity features for investigating networks in health and disease. 
A central challenge in this effort is to match the individual-level spatial maps to a population template in order to establish correspondences across subjects. 
Common approaches to obtain subject-specific networks with group correspondence often incorporate back-projection and dual regression ~\cite{beckmann2009dual, pmid11559959}, or hierarchical priors within unsupervised learning \cite{varoquaux2011msdl, kong2018}.  
While a number of studies have developed subject-specific parcellations, the significance of this inter-subject variability for network analysis has only recently been discussed. 
Kong et al. \cite{kong2018} developed high quality subject-specific parcellations using a multi-session hierarchical Bayesian model, and showed that subject-specific variability in functional topography can predict behavioral measures.  
Recently, using a novel parcellation scheme based on K-medoids clustering, Salehi et al. \cite{Salehi2018} showed that individual-level parcellation alone can predict the sex of the individual. 
These studies suggest the intriguing idea that subject-level network organization,  i.e. voxel-to-network assignments, can capture concepts intrinsic to individuals, just like connectivity strength. 
%

\paragraph{III. Is there a universal 'gold-standard' atlas? }

When considering the family of different methods, algorithms or modalities , there exist a plethora of  diverse brain parcellations at varying levels of granularity. Thus far, there is no unified framework for reasoning about these brain parcellations. Several taxonomic classifications can be used to describe the generation of these parcellations, such as machine learning or boundary detection, decomposition or clustering, multi-modal or unimodal. Even within the large class of clustering approaches, it is impossible to find a single algorithm that is consistently superior for a collection of simple, desired properties of partitioning \cite{Kleinberg2002}. Several evaluation criteria have emerged for comparing different parcellations, exposing the inherent trade-offs at work. Arslan et al.~\cite{arslan2018} performed an extensive comparison of several parcellations across diverse methods on resting-data from the Human Connectome Project (HCP). Through independent evaluations, they concluded that no single parcellation is consistently superior across all evaluation metrics.
Recently, Salehi et al. \cite{Salehi18fc} showed that different functional conditions, such as task or rest, generate reproducibly distinct parcellations thus questioning the very existence of an optimal parcellation, even at an individual-level. These novel studies necessitate rethinking about the final goals of brain mapping. Several studies have reflected the view that there is no optimal functional division of the brain, rather just an array of meaningful brain parcellations \cite{Blumensath2013}. Perhaps, brain mapping should not aim to identify functional sub-units in a universal sense, like Broadmann areas. Rather, the goal of human brain mapping should be reformulated as revealing consistent functional delineations that enable reliable and meaningful investigations into brain networks.

\paragraph{IV. A comparison between decomposition and clustering}

A high degree of convergence has been observed in the functionally coherent patterns extracted using decomposition and clustering.  Decomposition techniques allow soft partitioning of the data, and can thus yield spatially overlapping networks. These models may be more natural representations of brain networks where, for example, highly integrated regions such as network ‘hubs’ can simultaneously subserve multiple functional systems. Although it is possible to threshold and relabel the generated maps to produce spatially contiguous brain parcellations, these techniques are not naturally designed to generate disjoint partitions. In contrast, clustering techniques automatically yield hard assignments of voxels to different brain networks. Spatial constraints can be easily incorporated within different clustering algorithms to yield contiguous parcels. Decomposition models can adapt to varying data distributions, whereas clustering solutions allow much less flexibility owing to rigid clustering objectives. For example, k-means clustering function looks to capture spherical clusters. While a thorough comparison between these approaches is still lacking, some studies have identified the trade-offs between choosing either technique for parcellation. Abraham et al. \cite{abraham2013tv} compared clustering approaches with group-ICA and dictionary learning on two evaluation metrics: stability as reflected by reproducibility in voxel assignments on independent data, and data fidelity captured by the explained variance on independent data.  They observed a stability-fidelity trade-off: while clustering models yield stable regions but do not explain test data as well, linear decomposition models explain the test data reasonably well but at the expense of reduced stability. 

\subsection{Discovering patterns of dynamic functional connectivity \label{app2}}

Unsupervised learning has also been applied to study patterns of temporal organization or dynamic reconfigurations in resting-state networks. These studies are often based on two alternate hypothesis that (a) dynamic (windowed) functional connectivity cycles between discrete "connectivity states", or (b) functional connectivity at any time can be expressed as a combination of latent "connectivity states".  The first hypothesis is examined using clustering-based approaches or generative models like HMMs, while the second is modelled using decomposition techniques. Once stable states are determined across population, the former approach allows us to estimate the fraction of time spent in each state by all subjects. This quantity, known as dwell time or occupancy of the state, shows meaningful variation across individuals \cite{Vidaurre2017,Allen2014, damaraju2014,rashid2014}.  It is important to note than in all these approaches, the RSNs or the spatial patterns are assumed to be stationary over time and it is the temporal coherence that changes with time. 



\begin{figure*}
\hspace*{-1cm}
\centering
  \includegraphics[width=17cm,keepaspectratio]{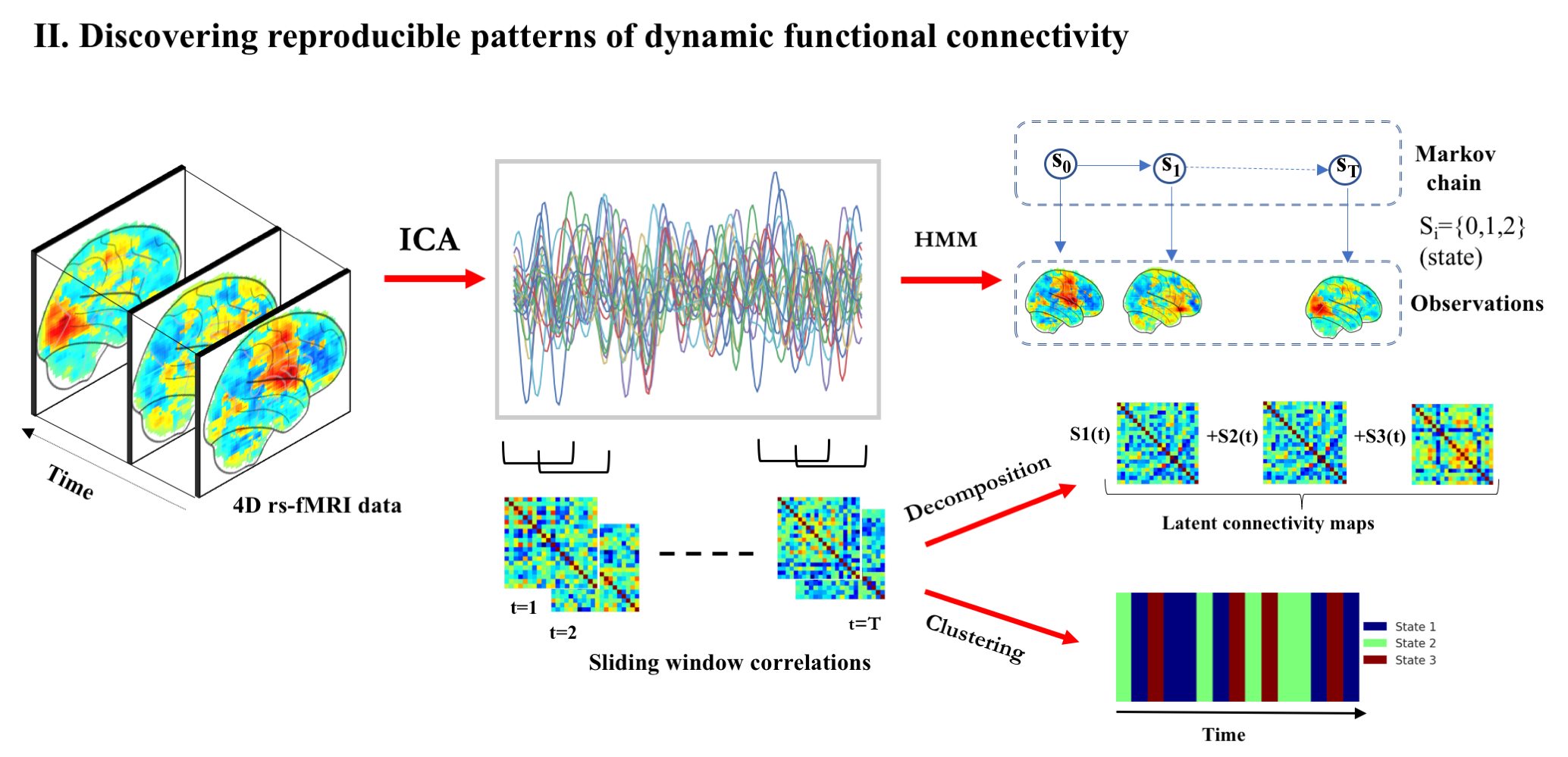}
  \caption{Schematic of application \ref{app2}. Three connectivity states are assumed in the data for illustration purposes}
\end{figure*}

\subsubsection{Clustering} 
Several studies have discovered recurring dynamic functional connectivity patterns, known as "states",  through k-means clustering of windowed correlation matrices \cite{Allen2014, damaraju2014, rashid2014, barber2018,abrol2017}. FC associated with these repeating states shows marked departure from static FC, suggesting that network dynamics provide novel signatures of the resting brain \cite{Allen2014}.  Notable differences have been observed in the dwell times of multiple states between healthy controls and patient populations across schizophrenia, bipolar disorder and psychotic-like experience domains \cite{damaraju2014,rashid2014, barber2018}.



Abrol et al. \cite{abrol2017} performed a large-scale study to characterize the replicability of brain states using standard k-means as well as a more flexible, soft k-means algorithm for state estimation. Experiments indicated reproducibility of most states, as well as their summary measures, such as mean dwell times and transition probabilities etc. across independent population samples.  While these studies establish the existence of recurring FC states, behavioral associations of these states is still unknown. In an interesting piece of work, Wang et al. \cite{wang2016dyn} identified two stable dynamic FC states using k-means clustering that showed correspondence with internal states of high- and low-arousal respectively. This suggests that RSFC fluctuations are behavioral state-dependent, and presents one explanation to account for the heterogeneity and dynamic nature of RSFC.


\subsubsection{Markov modelling of state transition dynamics} 
Hidden Markov Models (HMMs) are a class of unsupervised learning methods for sequential data. They are used to model a Markov process where states are not observed; what is observed are entities generated by these discrete states. HMMs are another valuable tool to interrogate recurring functional connectivity patterns \cite{Eavani2013, Vidaurre2017, suk2016}. The notion of states remains similar to the "FC states" described above for clustering; however, the characterization and estimation is drastically different. Unlike clustering where sliding windows are used to compute dynamic FC patterns, HMMs model the rs-fMRI time-series directly. Hence, they offer a promising alternative to overcome statistical limitations of sliding-windows in characterizing FC changes.


Several interesting results have emerged through the adoption of HMMs. Vidaurre et al. \cite{Vidaurre2017} find that relative occupancy of different states is a subject-specific measure linked with behavioral traits and heredity. Through Markov modelling, transitions between states have been revealed to occur as a non-random sequence \cite{Allen2014, Vidaurre2017}, that is itself hierarchically organized \cite{Vidaurre2017}. Recently, network dynamics modelled using HMMs were shown to distinguish MCI patients from controls \cite{suk2016}, thereby indicating their utility in clinical domains. 



 \subsubsection{Finding latent connectivity patterns across time-points}
 Decomposition techniques for understanding RSFC dynamics have the same flavor as the ones described in section -- of explaining data through latent factors; however, the variation of interest is across time in this case. Adoption of  matrix decomposition techniques exposes a basis set of FC patterns from windowed correlation matrices. Dynamic FC has been characterized using varied decomposition approaches, including PCA\cite{Leonardi2013}, Singular Value Decomposition (SVD)\cite{Leonardi2014}, non-negative matrix factorization\cite{chai2017} and sparse dictionary learning\cite{li2014ptsd}. 


Decomposition approaches, here, diverge from clustering or HMMs as they associate each dFC matrix with multiple latent factors instead of a single component. To compare these alternate approaches, Leonardi et al. \cite{Leonardi2014} implemented a generalized matrix decomposition, termed k-SVD. This factorization generalizes both k-means clustering and PCA subject to variable constraints. Reproducibility analysis in this study indicated that dFC is better characterized by multiple overlapping FC patterns.  

Decomposition of dFC has revealed novel alterations in network dynamics between healthy controls and patients suffering from PTSD \cite{li2014ptsd} or multiple sclerosis \cite{Leonardi2013}, as well as between childhood and young adulthood \cite{chai2017}. 

\subsection{Disentangling latent factors of inter-subject FC variation\label{app3}} 
Unsupervised learning can also disentangle latent explanatory factors for FC variation across population. We find two applications here: (i) learning low dimensional embeddings of FC matrices for subsequent supervised learning and (ii) learning population groupings to differentiate phenotypes based solely on FC. 

\subsubsection{Dimensionality reduction}
Rs-fMRI analysis is plagued by the {curse of dimensionality}, i.e., the phenomenon of increasing data sparsity in higher dimensions. Commonly used data features such as FC between pairs of regions, increase as $\mathcal{O}(n^2)$ with the number of parcellated regions. Further, sample size in typical fMRI studies is typically of the order of tens or hundreds, making it harder to learn generalizable patterns from original high dimensional data. To overcome this, linear decomposition methods like PCA or sparse dictionary learning have been widely used for dimensionality reduction of functional connectivity data \cite{calhoun2011psych,amico2018,eavani2015sparse,eavani2014miccai}.

Several non-linear embedding methods like Locally linear embedding (LLE) or Autoencoders (AEs)  have also garnered attention. LLE projects data to a reduced dimensional space while preserving local distances between data points and their neighborhood.  LLE embeddings have been employed in rs-fMRI studies, for example, to improve predictions in supervised age regression~\cite{Qiu2015}, or for low-dimensional clustering to distinguish Schizophrenia patients from controls~\cite{shen2010lle}. AEs are a neural network based alternative for generating reduced feature sets through nonlinear input transformations. Through an encoder-decoder style architecture, AEs are designed to capture compressed input representations that approximate inputs with minimal reconstruction loss. They have been used for feature reduction of RSFC in several studies \cite{suk2016, guo2017} . AEs can also be used in a pre-training stage for supervised neural network training, in order to direct the learning towards parameter spaces that support generalization \cite{Erhan2010}. This technique was shown, for example, to improve classification performance of Autism and Schizophrenia using RSFC~\cite{heinsfeld2018,Kim2016scz}.  



\begin{figure}
\centering
  \includegraphics[width=8cm,keepaspectratio]{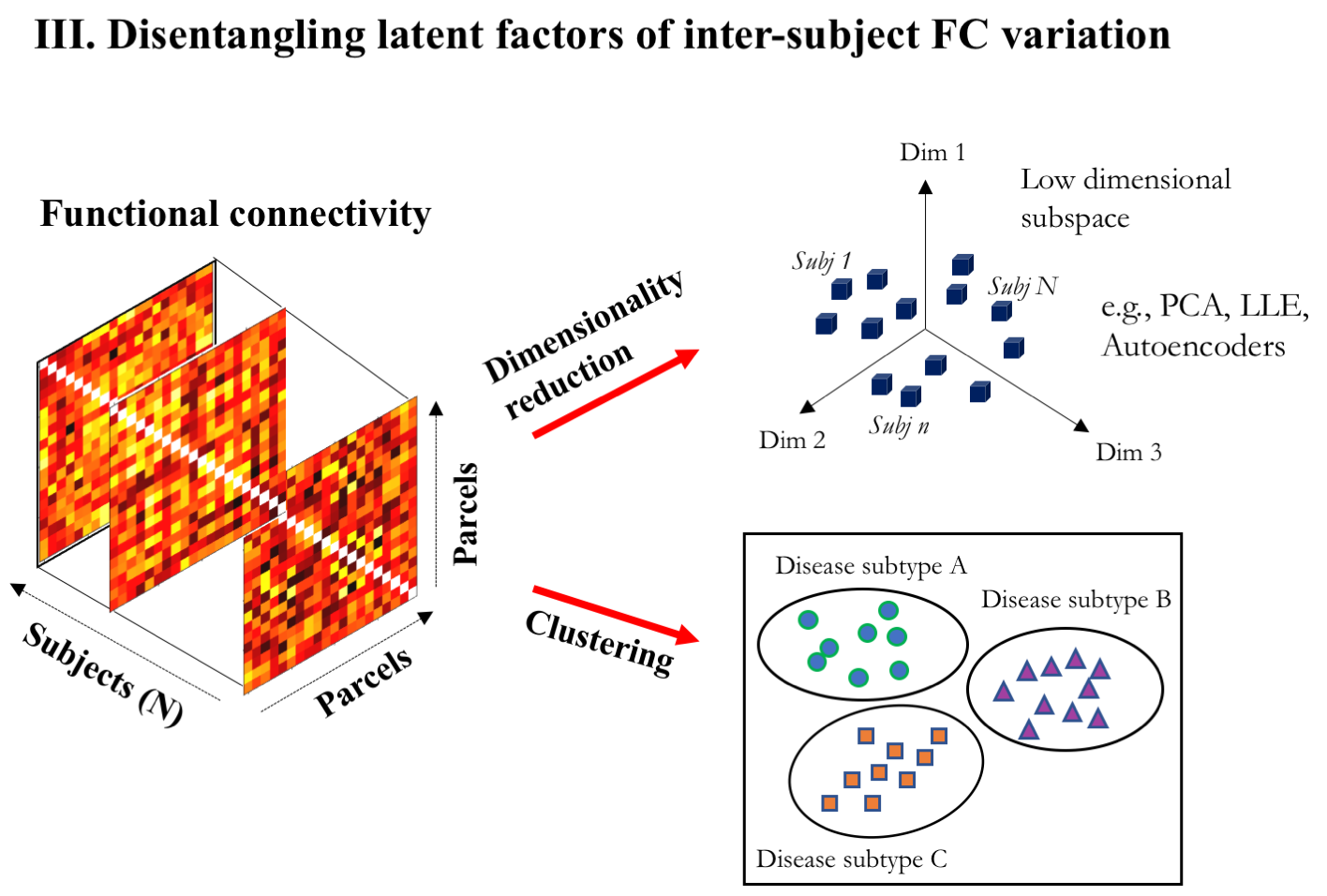}
  \caption{Schematic of application \ref{app3}. Dimensionality reduction of high-dimensional connectomes into 3 latent components is shown for illustration.}
\end{figure}

\subsubsection{Clustering heterogeneous diseases}
Clustering can expose sub-groups within a population that show similar FC.  Using unsupervised maximum margin clustering~\cite{NIPS2004_2602}, Zeng et al.~\cite{zeng2014} demonstrated that clusters can be associated with disease category (depressed v/s control) to yield high classification accuracy. Recently, Drysdale et al. \cite{drysdale2017dep}  discovered novel neurophysiological subtypes of depression based on RSFC. Using an agglomerative hierarchical procedure, they identified clustered patterns of dysfunctional connectivity, where clusters showed associations with distinct clinical symptom profiles despite no external supervision. Several psychiatric disorders, like depression, schizophrenia, and autism spectrum disorder, are believed to be highly heterogeneous with widely varying clinical presentations. Instead of labelling them as a unitary syndrome, differential characterization based on disease sub-types can build better diagnostic, prognostic or therapy selection systems. Unsupervised clustering could aid in the identification of these disease subtypes based on their rs-fMRI manifestations.



    

\section{Supervised Learning}
Supervised learning denotes the class of problems where the learning system is provided input features of the data and corresponding target predictions (or labels). 
The goal is to learn the mapping between input and label, so that the system can compute predictions for previously unseen input data points. 
Prediction of autism from rs-fMRI correlations is an example problem. Since intrinsic FC reflects interactions between cognitively associated functional networks, it is hypothesized that systematic alterations in resting-state patterns can be associated with pathology or cognitive traits. Promising diagnostic accuracy attained by supervised algorithms using rs-fMRI constitute strong evidence for this hypothesis. 




In this section, we separate the discussion of rs-fMRI feature extraction from the classification algorithms and application domains. 
\begin{figure*}
\hspace*{-0.8cm}
\centering
  \includegraphics[width=\textwidth,keepaspectratio]{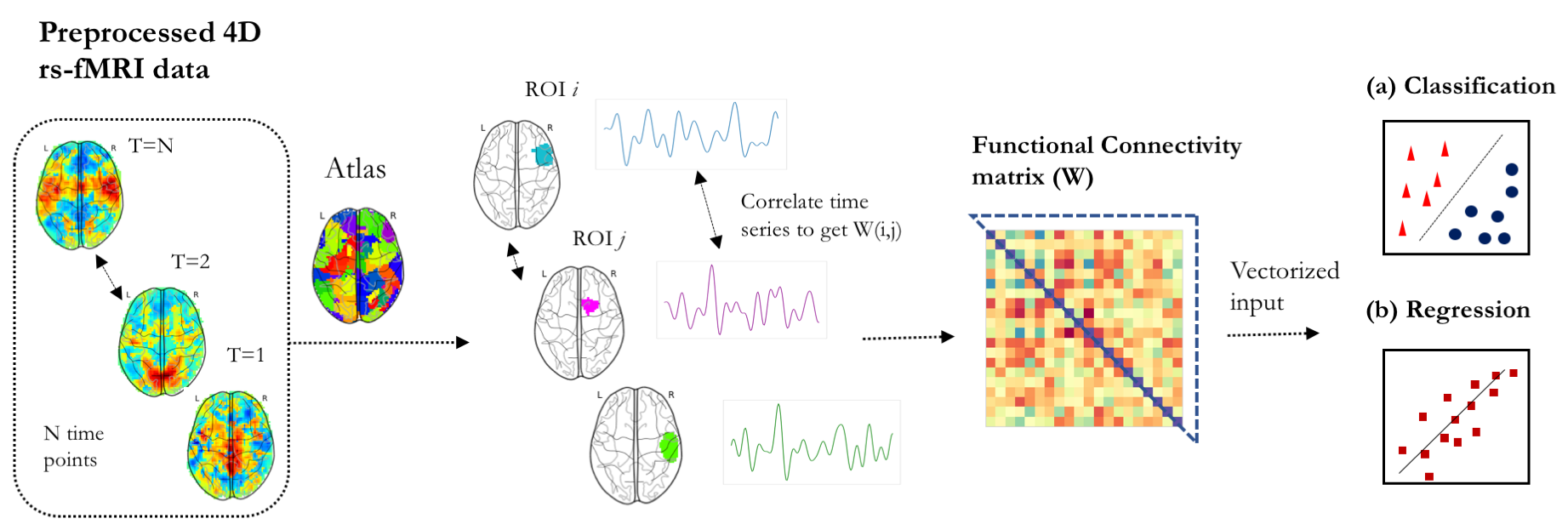}
  \caption{A common classification/regression pipeline for connectomes}
  \label{fig:supervised}
\end{figure*}

\subsection{Deriving connectomic features} 
To render supervised learning effective, the most critical factor is feature extraction. Capturing relevant neurophenotypes from rs-fMRI depends on various design choices. Almost all supervised prediction models use brain networks or "connectomes" extracted from rs-fMRI time-series as input features for the learning algorithm. The prototypical prediction pipeline is shown in Figure \ref{fig:supervised}. Here, we discuss critical aspects of common choices for brain network representations in supervised learning.   

The first step in the prototypical pipeline is region definition and corresponding time-series extraction. Dense connectomes derived from voxel-level correlations are rarely used in practice for supervised prediction due to their high dimensionality.  Both functional and anatomical atlases have been extensively used for this dimensionality reduction. Atlases delineate ROIs within the brain that are often used to study RSFC at a supervoxel scale. Each ROI is represented with a distinct time-course, often computed as the average signal from all voxels within the ROI. Consequently, the data is represented as an $N \times T$ matrix, where $N$ denotes the number of ROIs and $T$ represents the time-points in the signal. A drawback of using pre-defined atlases is that they may not explain the rs-fMRI dataset very well since they are not optimized for the data at hand. Several studies employ data-driven techniques to define regions within the brain, using unsupervised models such as K-means clustering, Ward clustering, ICA or dictionary learning etc \cite{Abraham2017SL, dadi2018}. It is important to note that since we use pairs of ROIs to define whole-brain RSFC, the features grow as $\mathcal{O}(N^2)$ with the number of ROIs. Therefore, in most studies, the network granularity is often limited to the range of 10-400 ROIs. 

The second step in this pipeline involves defining connectivity strength for extracting the connectome matrix. Functional connectivity between pairs of ROIs is the most common feature representation of rs-fMRI in supervised learning. In order to extract connectivity matrix, first the covariance matrix needs to be estimated. 
Sample covariance matrices are subject to a significant amount of estimation error due to the limited number of time-points. 
This ill-posed problem can be partially resolved through the use of shrinkage transformations~\cite{Varoquaux2010covariance}. Connectivity strength can then be estimated from the covariance matrix in multiple ways. 
Pearson's correlation coefficient is a commonly used metric for estimating functional connectivity. 
Partial correlation is another metric that has been shown to yield better estimates of network connections in simulated rs-fMRI data~\cite{Smith2011}. 
It measures the normalized correlation between two time-series, after removing the effect of all other time-series in the data. 
Alternatively, one can use a tangent-based reparametrization of the covariance matrix to obtain functional connectivity matrices that respect the Riemannian manifold of covariance matrices~\cite{Varoquaux2010stroke}. 
These connectivity coefficients can boost the sensitivity for comparing diseased versus patient populations~\cite{Abraham2017SL, Varoquaux2010stroke}. 
It is also possible to define frequency-specific connectivity strength by decomposing the original time-series into multiple frequency sub-bands and correlating signals separately within these sub-bands~\cite{Richiardi2011}.  

A few studies depart from this routine. In graph-theoretic analysis, it is common to represent parcellated brain regions as graph nodes and functional connectivity between nodes as edge weights. This graph based representation of functional connectivity, the human "connectome",  has been used to infer various topological characteristics of brain networks, such as modularity, clustering, small-worldedness etc. Some discriminative models have exploited these graph-based measures for individual-level predictions \cite{Khazaee2015alz, Lord2012, supekar2008}, although they are more commonly used for comparing groups. While limited in number, a few studies have also explored rs-fMRI features beyond RSFC. Amplitude of low-frequency fluctuations (ALFF) and local synchronization of rs-fMRI signals or Regional Homeogeneity (ReHo) are two alternate measures for studying spontaneous brain activity that have shown discriminative ability~\cite{zhu2005reho, Mennes2011alff}. 
More recently, several studies have also begun to explore the predictive capacity of dynamic FC in supervised models~\cite{Price2014, Madhyastha2015}.


\subsection{Algorithms}

The majority of supervised learning methods applied to rs-fMRI are discriminant-based, i.e., they discriminate between classes without any prior assumptions about the generative process, thereby bypassing the estimation of likelihoods and posterior densities. The focus is on correctly estimating the boundaries between classes of interest. Learning algorithms for the same discriminant function (e.g., linear) can be based on different objective functions, giving rise to distinct models. We describe common models below.

\subsubsection{Support Vector Machines (SVMs) and kernels}

The Support vector machine is the most widely used classification/regression algorithm in rs-fMRI studies~\cite{orrureview}. 
SVMs search for an optimal separating hyperplane between classes that maximizes the margin, i.e., the distance from hyperplane to points closest to it on either side~\cite{Cortes1995}. 
The resulting classification model is determined by only a subset of training instances that are closest to the boundary, known as the support vectors. 
SVMs can be extended to seek non-linear separating boundaries via adopting a so-called kernel function. 
The kernel function, which quantifies the similarity between pairs of points, implicitly maps the input data to higher dimensions. Conceptually, the use of kernel functions allows incorporation of domain-specific measures of similarity. 
For example, graph-based kernels can define a distance metric on the graphical representation of functional connectivity data for classification directly in the graph space~\cite{Takerkart2014}.  


\subsubsection{Neural Networks}
An ideal machine learning system should be highly automated, with limited hand-crafting in feature extraction as well as minimal assumptions about the nature of mapping between data and labels. 
The system should be able to mechanistically learn patterns useful for prediction from observed labelled data. 
Neural networks are highly promising methods for automated learning. This stems from their capability to approximate arbitrarily complex functions given sufficient labelled data~\cite{HORNIK1989359}. Traditionally, the use of neural network algorithms has been limited since neuroimaging is a data-scarce domain, making it difficult to learn a reliable mapping between input and prediction variables. However, with data sharing and open release of large-scale neuroimaging data repositories, neural networks have recently gained adoption in the the rs-fMRI community for supervised prediction tasks. Loosely inspired by the brain's architecture, neural networks comprise layers of feature extraction units that learn multiple levels of abstraction from the data directly. 
Neural networks with fully connected dense layers have been adopted to learn arbitrary mappings from connectivity features to disease labels~\cite{heinsfeld2018,Kim2016scz}. 
Recently, more advanced neural networks models with local receptive fields, like convolutional neural networks (CNNs), have shown promising classification accuracy using rs-fMRI data~\cite{Khosla2018}. 
Success of this approach stems from its ability to exploit the full-resolution 3D spatial structure
of rs-fMRI without having to learn too many model parameters, thanks to the weight sharing in CNNs.  








\subsection{Applications of supervised learning}
Studies harnessing resting-state correlations for supervised prediction tasks are evolving at an unprecedented scale. We describe some interesting applications of supervised machine learning in rs-fMRI below. 

 \subsubsection{ Brain development and aging}
 Machine learning methods have shown promise in investigating the developing connectome. In an early influential work, Dosenbach et al.~\cite{Dosenbach2010age} demonstrated the feasibility of using RSFC to predict brain maturation as measured by chronological age, in adolescents and young adults. Using SVR, they developed a functional maturation index based on predicted brain ages. Later studies showed that brain maturity can be reasonably predicted even in diverse cohorts distributed across the human lifespan \cite{Wang2012age,Meier2012}. These works posited rs-fMRI as a valuable tool to predict healthy neurodevelopment and exposed novel age-related dynamics of RSFC, such as major changes in FC of sensorimotor regions~\cite{Meier2012}, or an increasingly distributed functional architecture with age~\cite{Dosenbach2010age}. In addition to characterizing RSFC changes accompanying natural aging, machine learning has also been used to identify atypical neurodevelopment~\cite{Ball2016}.
 
 
 \subsubsection{Neurological and Psychiatric Disorders}
 Machine learning has been extensively deployed to investigate the diagnostic value of rs-fMRI data in various neurological and psychiatric conditions. Neurodegenerative diseases like Alzheimer's disease~\cite{chen2011alz,Khazaee2015alz, Challis2015}, its prodromal state Mild cognitive impairment \cite{Wee2014mci, chen2016, jie2014mci, wee2016mci}, Parkinson's \cite{Long2012pd}, and Amyotrophic Lateral Sclerosis (ALS) \cite{Welsh2013} have been classified by ML models with promising accuracy using functional connectivity-based biomarkers. 
 Brain atrophy patterns in neurological disorders like Alzheimer's or Multiple Sclerosis appear well before before behavioral symptoms emerge. Thus, neuroimaging-based biomarkers derived from structural or functional abnormalities are favorable for early diagnosis and subsequent intervention to slow down the degenerative process.
 
 The biological basis of psychiatric disorders has been elusive and the diagnosis of these disorders is currently completely driven by behavioral assessments. rs-fMRI has emerged as a powerful modality to derive imaging-based biomarkers for making diagnostic predictions of psychiatric disorders. Supervised learning algorithms using RSFC have shown promising results for classifying or predicting symptom severity in a variety of psychiatric disorders, including schizophrenia~\cite{ Venkataraman2012scz, Kim2016scz, bassett2012scz, Fan2011scz}, depression~\cite{cc2009dep, zeng2012,Lord2012}, autism spectrum disorder~\cite{Nielsen2013, Price2014, Abraham2017SL,Khosla2018}, attention-deficit hyperactivity disorder~\cite{eloyan2012,fair2012adhd}, social anxiety disorder~\cite{liu2015}, post-traumatic stress disorder~\cite{gong2014} and obsessive compulsive disorder~\cite{harrison2009}. Several novel network disruption hypotheses have emerged for these disorders as a consequence of these studies. Most of these prediction models are based on standard kernel-based SVMs, and rely on FC between ROI pairs as discriminative features. 
 
 \subsubsection{Cognitive abilities and personality traits}
 
 Functional connectivity can also be used to predict individual differences in cognition and behavior~\cite{mueller2013}.  In comparison to task-fMRI studies which capture a single cognitive dimension, the resting state encompasses a wide repertoire of cognitive states due to its uncontrolled nature. This makes it a rich modality to capture inter-individual variability across multiple behavioral domains. ML models have been shown to predict fluid intelligence~\cite{Finn2015}, sustained attention~\cite{rosenberg2016}, memory performance~\cite{siegel2016,meskaldji2016, jangraw2018}, language scores~\cite{siegel2016} from RSFC-based biomarkers in healthy and pathological populations. Recently, the utility of these models was also shown to extend to personality traits such as neuroticism, extraversion, agreeableness and openness~\cite{hsu2018, nostro2018}. 
 
 Prediction of behavioral performance is useful in a clinical context to understand how RSFC disruptions in pathology relate to impaired cognitive functioning. Meskaldji et al.~\cite{meskaldji2016} used regression models to predict memory impairment in MCI patients from different connectivity measures. Siegel et al.~\cite{siegel2016} assessed the behavioral significance of network disruptions in stroke patients by training ridge regression models to relate RSFC and structure with performance in multiple domains (memory, language, attention, visual and motor tasks). Among them, memory deficits were better predicted by RSFC, whereas structure was more important for predicting visual and motor impairments. This study highlights how rs-fMRI can complement structural information in studying brain-behavior relationships. 
 
 \subsubsection{Vigilance fluctuations and sleep studies}
 A handful of studies have employed machine learning to predict vigilance levels during rs-fMRI scans. Since resting-state studies demand no task-processing, subjects are prone to drifting between wakefulness and sleep. Classification of vigilance states during rs-fMRI is important to remove vigilance confounds and contamination. SVM classifiers trained on cortico-cortical RSFC have been shown to reliably detect periods of sleep within the sca~\cite{Tagliazucchi2012, Tagliazucchi2014}. Tagliazucchi et al. \cite{Tagliazucchi2014} revealed loss of wakefulness in one-third subjects of the experimental cohort, as early as 3 minutes into the scanner.  The findings are interesting: While resting state is assumed to capture wakefulness, this may not be entirely true even for very short scan durations. 
 The utility of these studies should not remain limited to classification alone. Through appropriate interpretation and visualization techniques, machine learning can shed new light on the reconfiguration of functional organization as people drift into sleep.

Predicting individual differences in cognitive response after different sleep conditions (e.g. sleep deprivation) using machine learning analysis of rs-fMRI is another interesting research direction. There is significant interest in examining RSFC alterations following sleep deprivation~\cite{dai2015, zhu2016}. While statistical analysis has elucidated the functional reorganization characteristic of sleep deprivation, much remains to be understood about the FC patterns associated with inter-individual differences in vulnerability to sleep deprivation.  Yeo et al. \cite{yeo2015sleep} trained an SVM classifier on functional connectivity data in the well-rested state to distinguish subjects vulnerable to vigilance decline following sleep deprivation from more resilient subjects, and revealed important network differences between the groups. 
 
 \subsubsection{Heritability}
Understanding the genetic influence on brain structure and function has been a long-standing goal in neuroscience. In a recent study, Ge et al. employed a traditional statistical framework to quantify heritability of whole-brain FC estimates~\cite{ ge2017}. Investigations into the genetic and environmental underpinnings of RSFC were also pursued within a machine learning framework. Miranda-Dominguez et al.~\cite{dominguez2018} trained an SVM classifier on individual FC signatures to distinguish sibling and twin pairs from unrelated subject pairs. The study unveiled several interesting findings. The ability to successfully predict familial relationships from resting-state fMRI indicates that aspects of functional connectivity are shaped by genetic or unique environmental factors. The fact that predictions remained accurate in young adult pairs suggests that these influences are sustained through development. Further, a higher accuracy of predicting twins compared to non-twin siblings implied that genetics (rather than environment) is likely the stronger predictive force. 
 
 \subsubsection{Other neuroimaging modalities}
 Machine learning can also be used to interrogate the correspondence between rs-fMRI and other modalities. The most closely related modality is task-fMRI.
Tavor et al. \cite{tavor2016} trained multiple regression models to show that resting-state connectivity can predict task-evoked responses in the brain across several behavioral domains. The ability of rs-fMRI, that is a task-free regime, to predict the activation pattern evoked by multiple tasks suggests that resting-state can capture the rich repertoire of cognitive states that is reflected during task-based fMRI. 
The performance of these regression models was shown to generalize to pathological populations~\cite{jones2017}, suggesting the clinical utility of this approach to map functional regions in populations incapable of performing certain tasks.  
 
 Investigating how structural connections shape functional associations between different brain regions has been the focus of a large number of studies~\cite{abdelnour2014}. While neuro-computational models have been promising to achieve this goal, machine learning models are particularly well-equipped to capture inter-individual differences in the structure-function relationship. Deligianni et al.~\cite{Deligianni2011} proposed a structured-output multivariate regression model to predict resting-state functional connectivity from DWI-derived structural connectivity, and demonstrated the efficiency of this technique through cross-validation. Venkataraman et al.~\cite{Venkataraman2012dwi} introduced a novel probabilistic model to examine the relationships between anatomical connectivity measured using DWI tractography and RSFC. Their formulation assumes that the two modalities are generated from a common connectivity template. Estimated latent connectivity estimates were shown to discriminate between control and schizophrenic populations, thereby indicating that joint modelling can also be useful in a clinical context.


\section{Discussion}

Many state-of-the-art techniques for rs-fMRI analysis are rooted in machine learning. Both unsupervised and supervised learning methods have substantially expanded the application domains of rs-fMRI. With large-scale compilation of neuroimaging data and progresses in learning algorithms, an even greater influence is expected in future. Despite the practical successes of machine learning, it is important to understand the challenges encountered in its current application to rs-fMRI. We outline some important limitations and unexplored opportunities below.



One of the biggest challenges associated with unsupervised learning methods is that there is no ground truth for evaluation. There is no \textit{a priori} universal functional map of the brain to base comparisons between parcellation schemes. 
Further, whole-brain parcellations are often defined at different scales of functional organization, ranging from a few large-scale parcels to several hundreds of regions, making comparisons even more challenging. Although several evaluation criteria have been developed that account for this variability, no single learning algorithm has emerged to be consistently superior in all. Due to the trade-offs among diverse approaches, the choice of which parcellation to use as reference for network analysis is thus largely subjective. 

Unsupervised learning approaches for exploring network dynamics are similarly prone to subjectivity. 
Characterizing dynamic functional connectivity through discrete mental states is difficult, primarily because the repertoire of mental states is possibly infinite. While dFC states are thought to reflect different cognitive processes, it is challenging to obtain a behavioral correspondence for distinct states since resting-state is not externally probed. 
This again makes interpretations hard and prone to subjective bias. 
 Machine learning approaches in this direction have thus far relied on cluster statistics to fix the number of FC states. Non-parametric models (e.g. infinite HMMs) provide an unexplored, attractive framework as they adaptively determine the number of states based on the underlying data complexity. 

A significant challenge in single-subject prediction using rs-fMRI is posed by the fact that rs-fMRI features can be described in multiple ways. There is no recognized gold-standard atlas for time-series extraction, nor is there a consensus on the optimal connectivity metric. Further, even the fMRI preprocessing strategies can vary considerably. Exploration across this space is cumbersome, especially for advanced machine learning models like neural networks that are slow to train. An ideal system should be invariant to these choices. However, this is hardly the case for rs-fMRI where large deviations have been reported in prediction performance in relationship to these factors~\cite{Abraham2017SL}.    

Another challenge in training robust prediction systems on large populations stems from the heterogeneity of multi-site rs-fMRI data. Resting-state is easier to standardize across sites compared to task-based protocols since it does not rely on external stimuli. However, differences in acquisition protocols and scanner characteristics across sites still constitute a significant source of heterogeneity. Multi-site studies have shown little to no improvement in prediction accuracy compared to single-site studies, despite the larger sample sizes~\cite{Nielsen2013, Dansereau2017}. 
While it is possible to normalize out site effects from data, more advanced tools are needed in practice to mitigate this bias. 

High diagnostic accuracies achieved by supervised learning methods should be interpreted with caution. Several confounding variables can induce systematic biases in estimates of functional connectivity. For example, head motion is known to affect connectivity patterns in the default mode network and frontoparietal control network~\cite{vandijk2012}. Further, motion profiles also vary systematically between subgroups of interest, e.g., diseased patients often move more than healthy controls. 
Apart from generating spurious associations, this could affect the interpretability of supervised prediction studies. 
Independent statistical analysis is critical to rule out the effect of confounding variables on predictions, especially when these variables differ across the groups being explored.


Methodological innovations are needed to improve prediction accuracy to levels suitable for clinical translation.  Several factors make comparison of methods across studies tedious. Cross-validation is the most commonly employed strategy for reporting performance of ML models. However, small sizes (common in rs-fMRI studies) are shown to yield large error bars~\cite{pmid28655633}, indicating that data-splits can significantly impact performance.  Generalizability and interpretability should remain the key focus while developing predictive models on rs-fMRI data. These are critical attributes to achieve clinical translation of machine learning models. Uncertainty estimation is another challenge in any application of supervised learning; ideally, class assignments by any classification algorithm should be accompanied by an additional measure that reflects the uncertainty in predictions. This is especially important for clinical diagnosis, where it is important to know a reliability measure for individual predictions. 

Most existing studies focus on classifying a single disease versus controls. The ability of a diagnostic system to discriminate between multiple psychiatric disorders is much more useful in a clinical setting~\cite{wolfers2015}. Hence, there is a need to assess the efficacy of ML models for differential diagnosis.  Integrating rs-fMRI with complementary modalities like diffusion-weighted MRI can possibly yield even better neurophenotypes of disease, and is another challenging yet promising research proposition. 




\section{Conclusions}
We have presented a comprehensive overview of the current state-of-the-art of machine learning in rs-fMRI analysis. We have organized the vast literature on this topic based upon applications and techniques separately to enable researchers from both neuroimaging and machine learning communities to identify gaps in current practice.  

\section*{Acknowledgements}
This work was supported by NIH R01 grants (R01LM012719 and R01AG053949), the NSF NeuroNex grant 1707312, and NSF CAREER grant (1748377).

\newpage

\bibliography{Reviewpapers}
\bibliographystyle{elsarticle-num}



\end{document}